\relax

\documentclass[letterpaper]{article} 
\pdfoutput=1

\usepackage{aaai21}  
\usepackage{times}  
\usepackage{helvet} 
\usepackage{courier}  
\usepackage[hyphens]{url}  
\usepackage{graphicx} 
\usepackage{natbib}  
\usepackage{caption} 
\usepackage{amsmath}

\usepackage{booktabs}
\usepackage{times}
\usepackage{epsfig}
\usepackage{graphicx,subfigure}
\usepackage{amssymb}
\usepackage{bm}
\usepackage{makecell}
\usepackage{multirow}
\usepackage{wrapfig}
\usepackage{algorithm}
\usepackage{algorithmic}
\usepackage{adjustbox}
\usepackage{bbm}
\usepackage{epstopdf}
\usepackage{threeparttable}
\usepackage{tablefootnote}
\usepackage[normalem]{ulem}

\usepackage{color}
\usepackage[dvipsnames]{xcolor}

\DeclareMathAlphabet{\mathcal}{OMS}{cmsy}{m}{n}

\usepackage{amsmath}
\usepackage{amsthm}

\usepackage[switch]{lineno}
 
\newcommand{\name}{Monet}
 
\graphicspath{{figs/}}

\begin{document}

\title{Motion-guided Non-local Spatial-Temporal Network for Video Crowd Counting}

\author {
    Haoyue Bai,
    S.-H. Gary Chan \\
}
\affiliations {
    Department of Computer Science and Engineering\\
    The Hong Kong University of Science and Technology, Hong Kong, China \\
    \{hbaiaa, gchan\}@cse.ust.hk
}

\maketitle
\begin{abstract}
We study video crowd counting, which is to estimate the number of objects (people in this paper) in all the frames of a video sequence.
Previous work on crowd counting is mostly on still images.
There has been little work on how to properly extract and take advantage of
the spatial-temporal correlation between neighboring frames
in both short and long ranges to achieve high estimation accuracy for a video sequence.
In this work, we propose \name{}, a novel and highly accurate {\bf mo}tion-guided {\bf n}on-local spatial-temporal n{\bf et}work for video crowd counting.

\name{}
first takes people flow (motion information) as guidance to coarsely segment 
the regions of pixels where a person may be.
Given these regions, \name{} then uses a non-local spatial-temporal network 
to extract spatial-temporally both short and long-range contextual information. The whole network is finally trained end-to-end with a fused loss
to generate a high-quality density map.
Noting the scarcity and low quality (in terms of resolution and scene diversity) of the publicly available video crowd datasets,
we have collected and built a large-scale video crowd counting datasets, VidCrowd, to contribute to the community. VidCrowd contains 9,000 frames of high resolution ($2560 \times 1440$), with 1,150,239 head annotations captured in different scenes, crowd density and lighting in two cities. We have conducted extensive experiments on the challenging VideoCrowd and two public video crowd counting datasets: UCSD and Mall. Our approach achieves substantially better performance in terms of MAE and MSE as compared with other state-of-the-art approaches.
\end{abstract}

\section{Introduction}
\label{sec:intro}

\begin{figure}[t]
    \centering
    \includegraphics[width=0.47\textwidth]{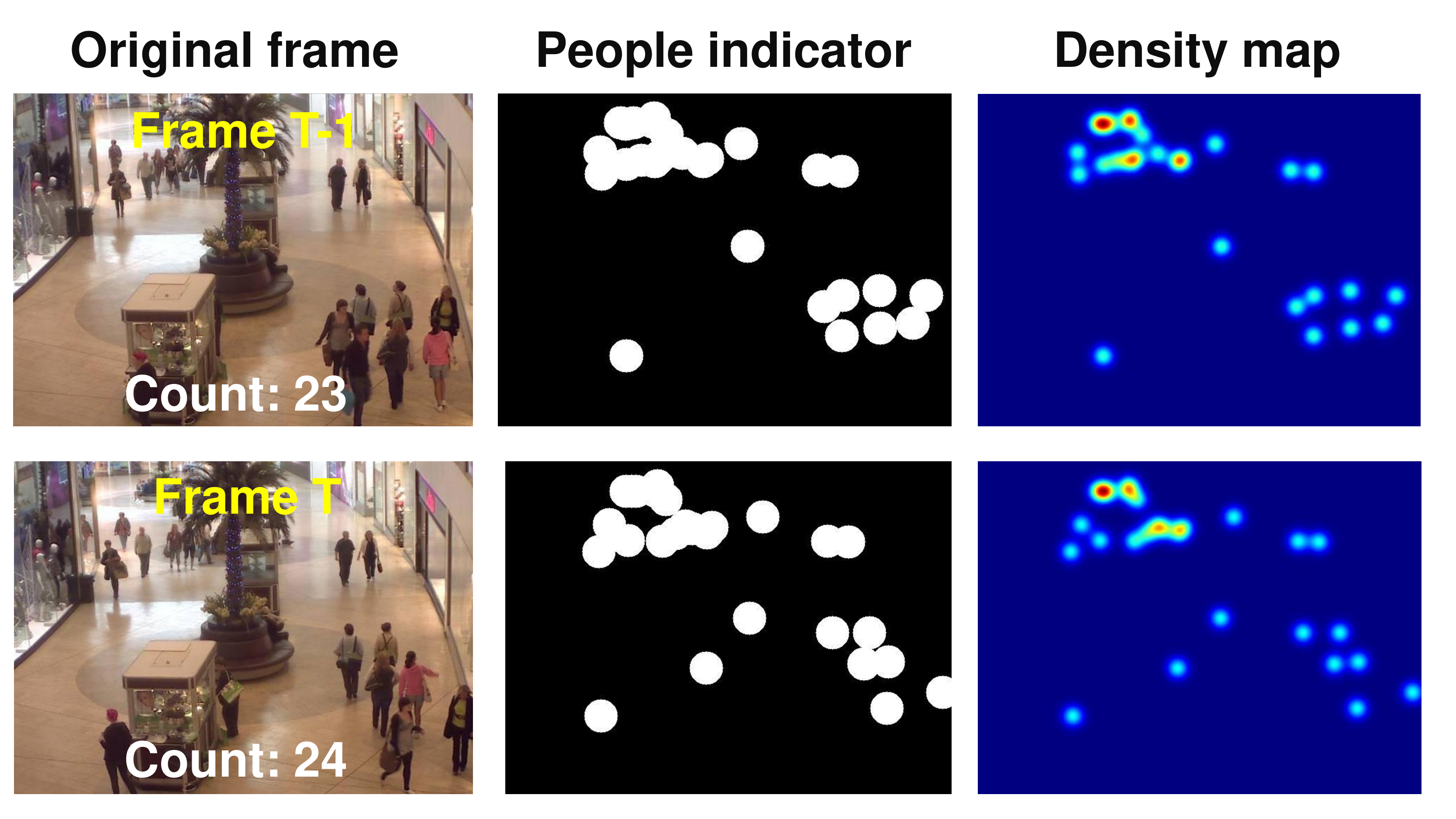}
    \caption{Crowd frames, areas with people and its corresponding density maps.}
    \label{image1}
\end{figure}

Crowd counting
is to estimate the number of objects (people in our case) in an image
of an unconstrained scene. It has attracted much attention due to its many applications in public safety, video surveillance, and traffic management~\cite{onoro2016towards, lempitsky2010learning, chan2008privacy, bai2020cnn}.
Counting in diverse real-world scenarios remains challenging due to severe occlusion, large scale variation, uneven distribution of people, etc.  
Recently, density map regression-based Convolutional Neural Networks (CNNs) have been extensively studied for crowd counting. Such approaches
incorporate spatial information to estimate the number of people per pixel 
in an image~\cite{pham2015count, lempitsky2010learning}.
It has shown to be promising with the incorporation of
multi-scale features, image pyramid architectures, special operations and attention mechanisms~\cite{zhang2016single, sam2017switching, cao2018scale, boominathan2016crowdnet}.

In this work, we investigate crowd counting in all the frames of a video sequence.
A straightforward but naive approach is to consider
the video frames independently by making use of the crowd counting techniques proposed before for still images. This is not satisfactory because it ignores the continuity or temporal correlation between frames, i.e., the motion information.
Bidirectional ConvLSTM is a recent attempt to exploit the correlation in video data~\cite{xiong2017spatiotemporal}. Despite encouraging results, the LSTM framework is not easy to train or to be extended to a general scenario. 
The 3D kernel is adopted to capture simultaneously the local temporal and spatial information.  While effective for slowly moving objects,  it is not effective in extracting the long-range contextual information, hence affecting its applicability in a general or fast-moving environment.
Notwithstanding the above, in video crowd counting research, another challenge is
the lack of large-scale publicly accessible datasets, mainly due to the cost in crowd data collection and annotation. As a result, the existing video crowd counting datasets only cover limited scene diversity, unsatisfactory resolution, and low crowd density.

To overcome the above challenges, we propose \name{},
a novel {\bf mo}tion-guided {\bf n}on-local spatial-temporal n{\bf et}work which captures both short and long-range spatial-temporal correlations between neighboring frames to achieve highly accurate video crowd counting. The crux of \name{} consists of three steps: 

\begin{enumerate}
	\item The motion estimation step, which computes people flow so as to segment the frame into coarse regions, consisting of groups of pixels where a person may be. Noting that people motion is usually different from the background motion, 
	this kind of motion vectors can be regarded as useful prior which provides informative clues to predict the spatial distribution of people.
	\item A non-local spatial-temporal network, which,
	guided by the segmented regions of the previous step,
	extracts both local (short-ranged) and non-local (long-ranged) context information from the consecutive frames to estimate the crowd.
	\item The motion guidance and the extracted non-local context information in space-time dimensions are integrated with cascaded refinement and a fused object function. Thus, the motion information can be effectively combined with the counting estimator and boosting the performance on video sequences. 
\end{enumerate}

In order to relieve the scarcity of the current
video crowd counting datasets and to 
enrich them
with a challenging one, we have built a new large-scale video crowd counting dataset, VidCrowd, to contribute to the community with more scene diversity, better resolution and higher crowd levels. VidCrowd dataset provides the community with 9,000 video frames of high resolution ($2560 \times 1440$) and 1,150,239 head annotations from two cities of 20 different scenes on campus, squares, park, street, beach, etc. 
VidCrowd also has diverse crowd levels and lighting conditions, and hence is a better candidate for video crowd counting evaluation.
In this paper, we conduct thorough and extensive experiments on the challenging vidcrowd dataset and two other public datasets (UCSD and Mall). 
\name{} outperforms exist video-based crowd counting methods on Mall and VidCrowd datasets in terms of MAE and MSE, and achieves comparable results with the state-of-the-art on UCSD dataset.

\begin{figure*}[!t]
	\centering
	\includegraphics[width=0.95\textwidth, height=0.39\textheight]{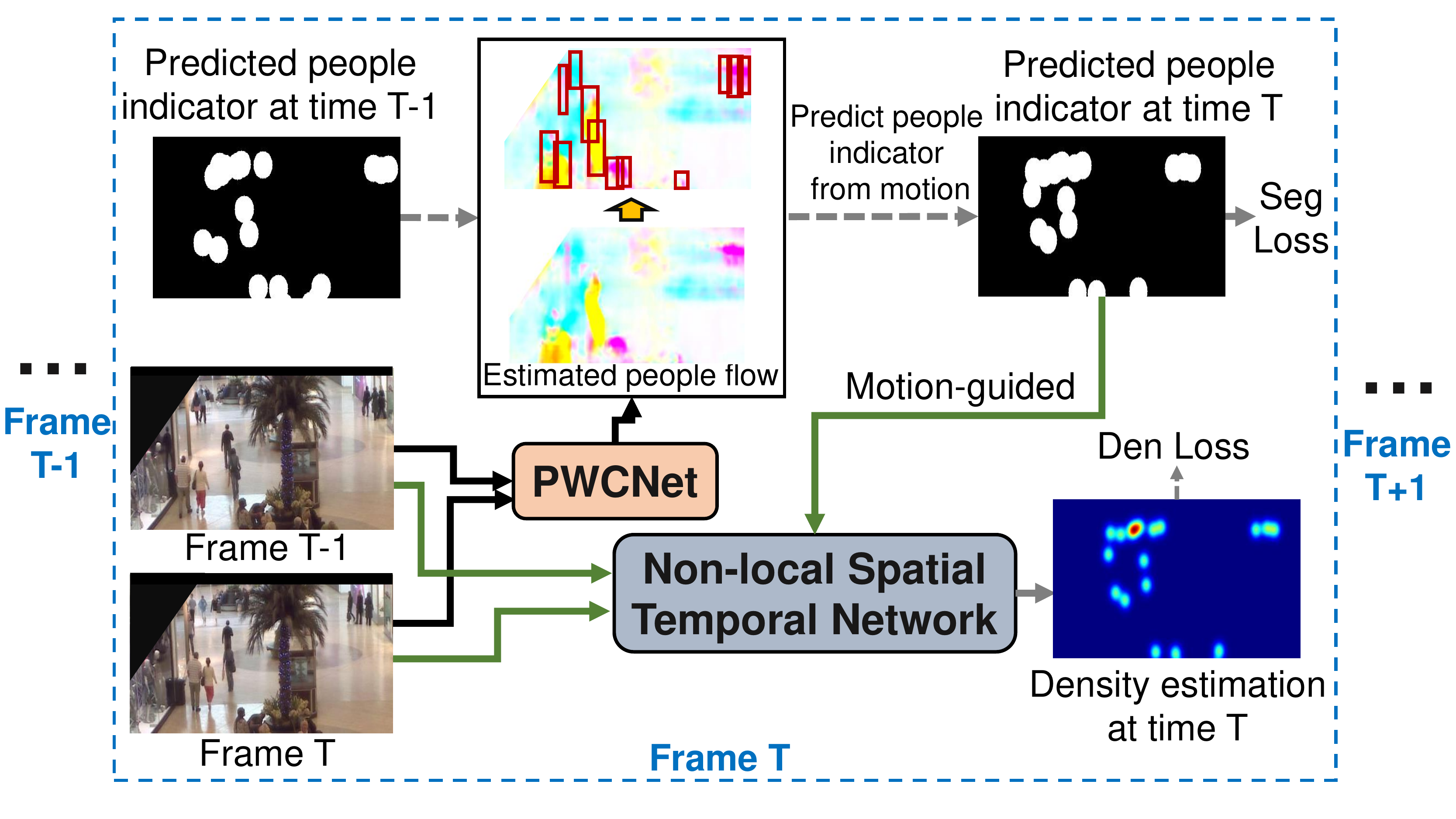}
	\caption{The framework of \name{}. 
	}
	\label{Monet}
\end{figure*}

This paper is organized as follows. We review related work
in Section~\ref{sect:related}, and
present the details of \name{}
in Section~\ref{sect:method}. We discuss our experimental setup and 
illustrative results in Section~\ref{sect:exp}, and 
conclude in Section~\ref{sect:conc}.

\section{Related Work} \label{sect:related}

In this section, we present the related work of crowd counting approaches in three main directions: traditional approaches (Section~\ref{tra}), deep learning-based approaches (Section~\ref{dl}), and crowd counting for video-based scenes (Section~\ref{video}).

\subsection{Traditional Approaches} \label{tra}
Early approaches for crowd counting are often based on detection models
, i.e., they leverage pedestrian or body-part detectors to detect individual objects and count the number
~\cite{rabaud2006counting, lin2010shape}. However, the performance of these works degrade quickly in highly crowded scenes. Some researchers have attempted to use regression-based approaches with low-level features like HOG and SIFT to calculate the global number~\cite{chan2012counting}. Even though relying on low-level features, these approaches achieve better results for the global count estimation. To incorporate spatial information, researchers have proposed the density map regression-based approaches, that is, measuring the number of people per unit pixel of an area in a crowd scene.  As discussed in~\cite{lempitsky2010learning}, the work is the first one to provide a density map regression-based crowd counting approach with linear mapping algorithms.  A subsequent work improves it with random forest regression to learn non-linear mapping and achieves much better performance~\cite{pham2015count}.

\subsection{Deep Learning-based Approaches}   \label{dl}
Recently, researchers have adopted deep learning-based methods instead of relying on hand-crafted features to generate high-quality density maps and achieve accurate crowd counting~\cite{cao2018scale, shen2018crowd, wang2020distribution, shi2020real}. These approaches can be applied to count different kinds of objects (i.e., vehicles and cells) instead of people~\cite{li2018csrnet, he2019automatic}.
Researchers propose multi-column convolutional neural networks with different kernel sizes for each column to address the scale variation problem~\cite{zhang2016single}. Switching-CNN attaches a patch-based switching block to the multi-column structure, and better handles the particular range of scale for each column~\cite{sam2017switching}. HydraCNN utilizes a pyramid of image patches with multiple scales for crowd estimation~\cite{onoro2016towards}. Some researchers also utilize image pyramid architectures, multi-scale features, and attention mechanisms to promote the counting performance~\cite{zhang2016single, cao2018scale, boominathan2016crowdnet, kang2018crowd, wang2019learning}.
However, the existing counting methods deal with each video frame independently, which will lose the strong temporal information hidden in motion.

\subsection{Video-based Crowd Counting}   \label{video}

Most of the previous works consider still image. There has been little work on video crowd counting where 
the correlation between consecutive frames should be considered~\cite{ren2020tracking, ma2021spatiotemporal}.
Bidirectional ConvLSTM is a recent approach to exploit the strong correlation in video frames~\cite{xiong2017spatiotemporal}. While encouraging, the LSTM module is hard to train and limits its wide applications to general scenarios.
3D kernel is utilized to extract temporal and spatial information simultaneously. While effective for extracting local features, E3D is not effective to extract the long-range correlations, thus hinder the performance~\cite{zou2019enhanced}.
Besides, the lack of large scale publicly available dataset is another challenge for video crowd counting research. Most of the existing video crowd counting datasets only cover limited scenes and with low crowd density due to the difficulties in crowd video data collection and annotation.
Compared with surveillance cameras to capture the crowd scenes, drone sensors are more flexible for smart city applications with larger coverage and higher resolution. Besides, compared with images taken by surveillance cameras, the isolated small clusters problem is more severe for drone-based crowd images, which brings more challenges for crowd estimation. Thus, we build a large-scale challenging video crowd counting datasets with 1,150,239 head annotations based on drone sensors to evaluate our algorithm and provide it to the community.


\section{\name{} Details}\label{sect:method}

In this section, we discuss the details of \name{}, a novel motion-guided non-local spatial-temporal network for video crowd counting.
In Section~\ref{meth:architecture}, we present an overview of the \name{} framework.
Section~\ref{meth:nonlocal} shows the details of the non-local spatial-temporal module. 
The objective function is described in Section~\ref{meth:object}. 
\subsection{Framework}
\label{meth:architecture}

\name{} captures the spatial and temporal correlations simultaneously for accurate video crowd counting. As shown in Fig.~\ref{Monet}, 
\name{} framework mainly contains three steps:
a) the motion estimation step is to compute people flow in order to segment the video frame into the coarse areas with people; b) guided by the segmented areas of the previous step, a non-local spatial-temporal network captures both local and non-local dependencies for crowd estimation; and c) the motion guidance 
and the extracted both short-range and long-range context information, are finally integrated in cascade to refine the estimated density maps with a fused objective function.

The people motion features are normally different from the background motion. Our target is the areas with people in a video frame, and the accuracy of estimation can be boosted if the influence of the background area can be reduced. \name{} takes in two consecutive frame T-1 and frame T and compute people flow for the input crowd scene. The estimated people flow map imposes strong constrains by coarsely segmenting the areas with people, and encoding the clues of the number of people.
As shown in the block of the estimated people flow in Fig~\ref{Monet}, 
different people moving speeds and directions are pixel-wisely color-encoded for visualization.
We also crop the individual objects with a red bounding box on the people flow map to better visualize the correspondence of the people flow map and the target areas with people.

Guided by the coarsely segmented areas of the first step, we propose a non-local spatial-temporal network to extract both local and non-local context information simultaneously and to combine the strong correlations between neighboring frames for accurate video crowd estimation. 
Within the non-local spatial-temporal module, the motion guidance vector and the spatial-temporal context information are integrated in a cascaded refinement manner towards a fused objective function.
The details of our non-local spatial-temporal module and the objective functions we used are discussed in the following sections.

\begin{figure*}[t]
	\centering
	\includegraphics[width=0.95\textwidth]{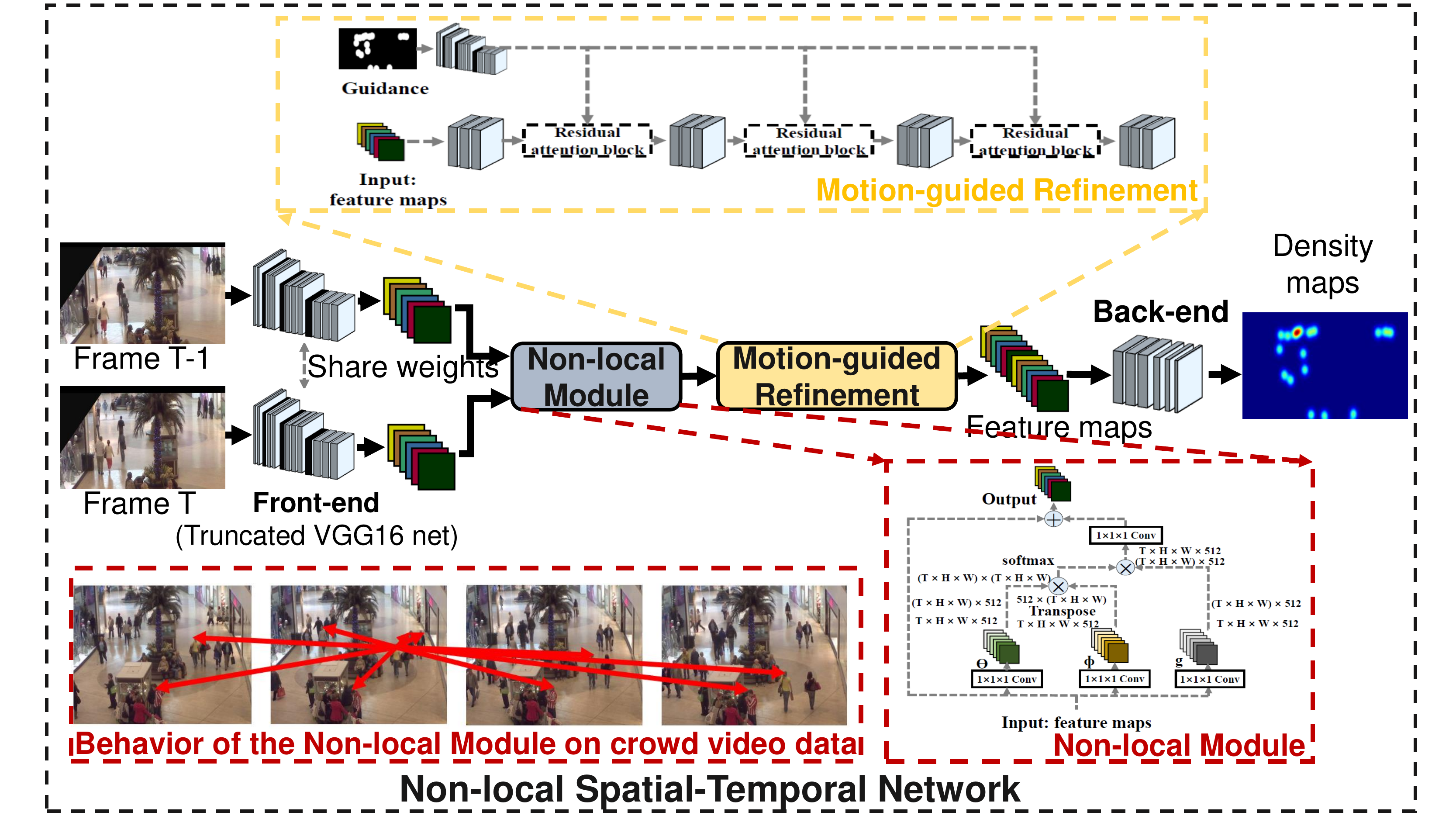}
	\caption{The details of the non-local spatial-temporal network.
	}
	\label{nonlocal}
\end{figure*}

\subsection{Non-local Spatial-Temporal Network} \label{meth:nonlocal}

The non-local spatial-temporal network, which guided by the coarse areas of people, is used to extract both the short-range and long-range context information for both the spatial and temporal wise simultaneously, and thus promote the performance for video crowd estimation. We present the details of our non-local spatial-temporal module in Fig.~\ref{nonlocal}. The non-local spatial-temporal network consists of four modules: front-end module, non-local module, motion-guided refinement, and back-end module.

The front-end module utilize the truncated VGG-16~\cite{simonyan2014very} with good transferability as the backbone for our \name{} for a fair comparison with the previous works~\cite{fang2019locality},~\cite{fang2020multi}.The first ten layers of VGG-16 with three pooling layers are extracted to balance the resolution and valid receptive field. As shown in the red box of Fig.~\ref{nonlocal}, we incorporate non-local operations into a non-local module in order to combine both local and non-local information from a video sequence. The general non-local operation~\cite{wang2018non} in a deep neural network can be defined as:
\begin{equation}
\begin{aligned}
\mathrm{y_i} = \frac{1}{C(x)}\sum_{\forall j}f(x_i,x_j)g(x_j),
\end{aligned}
\end{equation}
where $i$ is the index of an output position whose responsibility is to be computed, and $j$ is the index of all the possible locations. $x$ is the input features, and $y$ is the output video frame of the same size as input $x$. The function $f$ computes a relationship between $i$ and all $j$, and the function $g$ denotes an affinity of the input $x$ at position $j$. In our experiment, we set $(1/C(x))\sum_{\forall j}f(x_i,x_j)$ as softmax computation, and then we have: 
\begin{equation}
\mbox{y} = softmax(x^{T}W_{\theta}^{T}W_{\phi}x)g(x)
\end{equation}
as shown in the non-local module part.
We also present an example with consecutive frames for the behavior of the non-local module on crowd video sequence, refer to Fig.~\ref{nonlocal}. The starting point represents one $x_i$ and the ending points represent $x_j$. We incorporate both the spatial-wise and temporal-wise non-local modules and combine both non-local and local information for density estimation.

For the motion-guided refinement module, The segmented areas with people from the first step are cascaded fused with the non-local spatial-temporal network and refine the quality of the density maps. All the residual attention block share the same structure, which contains two inputs (input feature maps and guidance) and one output (refined feature maps). And this residual attention block is a variant of the residual block with spatial-wise and channel-wise attention. The motion-guided refinement allows the network to effectively combine with the guidance for accurate crowd estimation. Finally, the Back-end is used to fuse the extracted local and non-local information with guidance to predict high-quality density maps.

\begin{table*}[!t]
    \centering
	\caption{\small Statistics of the three labeled datasets for video crowd counting.}
	\begin{center}
		\label{table:dataset}
		\begin{adjustbox}{max width=1\textwidth}
		\begin{tabular}{lccccc}
			\toprule
			\toprule
			Dataset         &Resolution  &Number  &Average  &Total&GT Generation \\
			\midrule
			Mall         &480$\times$640   &2,000   &31.2 &62,315    &Geometry-adaptive \\
			UCSD         &158$\times$238   &2,000   &24.9 &49,885     &Fixed kernel:$\sigma=4$\\
			VidCrowd     &1440$\times$2560 &9,000   &127.8&1,150,239  &Fixed kernel:$\sigma=5$ \\
			\bottomrule
			\bottomrule
		\end{tabular} 
		\end{adjustbox}
	\end{center}
\end{table*}

\subsection{Objective Function}  \label{meth:object}   

The overall objective function is combined with the losses of the segmentation loss in the first step and the density map loss in the second step. The total loss function is defined as:
\begin{equation}
L_{total} = L_{den} + \lambda L_{seg},
\end{equation}
where $\lambda$ is balancing factors for the two losses.

To be specific, we use the same pixel-wise
Euclidean loss for density map regression~\cite{zhang2016single}. The Euclidean loss is defined as: 
\begin{equation}
L_{den}=\frac{1}{N} \left ||F(X;\alpha)-Y \right ||_{2}^{2},
\end{equation}
where $\alpha$ denotes the model parameters, N is the number of pixels, X presents the input image and Y represents the ground truth density map and $F(X;\alpha)$ is the predicted map.
The predicted counting results can be drawn from a sum over the predicted density map.

We use the BCE loss as the object function in the first stage to coarsely segment the areas with people. The BCE loss for segmentation is defined as
\begin{equation}
L_{seg}=-\frac{1}{N}\sum_{i=1}^{N}y_i\log(m_i) + (1-y_i)\log(1-m_i),
\end{equation}
where $y_i$ is the ground truth, $N$ is the number of samples, $m_i$ is the predicted coarse segmentation area with people. The ground truth of segmentation is generated from the original head annotation in our experiment.


\section{Experiments and Illustrative Results} \label{sect:exp}

In this section, we discuss the training details and evaluation metrics in Section~\ref{metrics}.
After that, we present illustrative results of \name{} on three different datasets: VidCrowd dataset (Section~\ref{vidcrowd}), Mall
and UCSD datasets (Section~\ref{mall}).

\subsection{Training Details and Evaluation Metrics} \label{metrics}

In the training stage, we randomly flip and crop the training video frames for data augmentation.
The optimization for the training stage is Adam solver, with a $10^{-5}$ learning rate, and the training batch size is 8 for all the three datasets in our experiment. To utilize the optical flow for the motion estimation module, we choose to use the pre-trained PWCNet~\cite{sun2018pwc} as described in Section~\ref{meth:architecture}. Our framework is implemented with PyTorch 1.1.0, CUDA v10.1. The code and the collected VidCrowd dataset will be released.

For the ground truth generation, we adopt the geometry-adaptive kernels to address the highly congested scenes for Mall dataset. The ground truth is generated by blurring each head annotation with a Gaussian kernel, which takes the spatial distribution of the video frame into considerations. The geometry-adaptive kernel is defined as follows:
$F(x) = \sum_{i=1}^{N}\delta(x-x_i) \times G_{\sigma_i}(x), with \; \sigma_i = \beta\bar{d_i}$,
where x denotes the pixel position in an image. For each target object, $x_i$ in the ground truth, which is presented with a delta function $\delta(x-x_i)$. The ground truth density map $F(x)$ is generated by convolving $\delta(x-x_i)$ with a normalized Gaussian kernel based on parameter $\sigma_i$.
And $\bar{d_i}$ shows the average distance of the k nearest neighbors.
As shown in Table~\ref{table:dataset}, We follow previous works to set $\beta = 0.3$ and $k = 3$ for Mall dataset and $k = 4$ for the UCSD dataset. For the Vidcrowd dataset, we use fixed kernel $\sigma$ = 5 as the ground truth generation method.

There are two metrics to evaluate the crowd counting results, Mean Absolute Error (MAE) and Mean Squared Error (MSE), which are defined as follows:
\begin{equation}
\begin{aligned}
\mathrm{MAE} = \frac{1}{N} \sum_{i=1}^{N}|C_i - \widehat{C}_i|,
\end{aligned}
\end{equation}

\begin{equation}
\begin{aligned}
\mathrm{MSE} = \sqrt{\frac{1}{N} \sum_{i=1}^{N}|C_i - \widehat{C}_i|^2},
\end{aligned}
\end{equation}
where $N$ presents the total number of test images, $C_{i}$ is the ground truth count of the $i$-th  input image, and $\hat{C}_{i}$ denotes the predicted counting results.

\begin{figure*}[!t]
	\centering
	\includegraphics[width=0.95\textwidth]{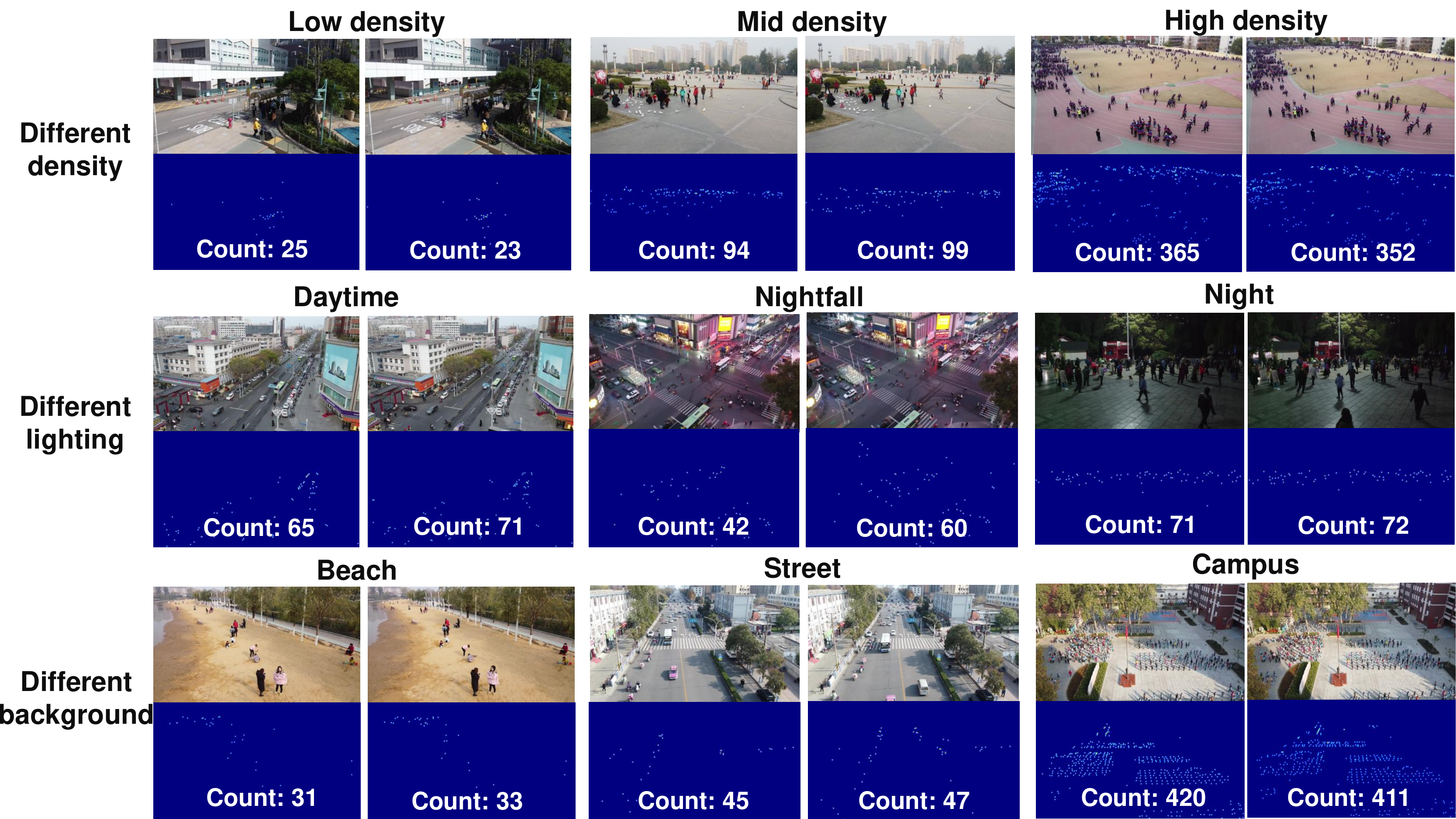}
	\caption{Visualization of our VidCrowd dataset.}
	\label{visual}
\end{figure*}

The comparison schemes in our experiments are below:

\begin{itemize}
	\item {\em ConvLSTM:} ConvLSTM~\cite{xiong2017spatiotemporal} based on a variant of Convolutional LSTM to incorporate both the spatial and temporal information and jointly combined to predict density maps.
	
	\item {\em E3D:} E3D~\cite{zou2019enhanced} utilize enhanced 3D convolutional into the counting networks for crowd counting, which is effective to extract local information and achieves superior counting performance.
	
	\item {\em LSTN:} LSTN~\cite{fang2019locality} and MLSTN~\cite{fang2020multi} leverage a kind of locality-constrained spatial transformer network to make use of temporal correlations for video-based crowd counting.
\end{itemize}

\begin{table*}[!t]
    \centering
	\caption{Results on VidCrowd dataset. All methods are implemented by ourselves.}
	\label{tab:vidcrowd}
	\begin{adjustbox}{max width=0.5\textwidth}
			\begin{tabular}{lcc}
				\toprule
				\toprule
				Method                  &MAE    &MSE     \\
				\midrule
				MCNN~\cite{zhang2016single}                    &29.49  &44.00   \\    
				CSRNet~\cite{li2018csrnet}                  &21.54  &36.72   \\
				SACANet~\cite{bai2019crowd}                 &18.52  &33.10  \\
				ConvLSTM~\cite{xiong2017spatiotemporal}                &17.23  &31.96   \\
				\hline
				\name{} (ours)          &$\textbf{15.06}$   &$\textbf{29.94}$    \\
				\bottomrule
				\bottomrule
			\end{tabular}
	\end{adjustbox}
\end{table*}

\subsection{Evaluation on VidCrowd Dataset} \label{vidcrowd}

We introduce a new large-scale video crowd counting dataset, VidCrowd, for the community. VidCrowd contains 9000 annotated video frames with 1,150,239 head annotations captured in different scenes and lighting across two cities. We use drone sensors to cover a larger area of scenes. The details of the VidCrowd dataset are presented in Table~\ref{table:dataset}. 
We collect the video data from both the daytime and the night to cover different lighting scenes in the real world.
The statistics of the three datasets are listed with the information of image resolution, the number of dataset instances, the average number of people for each image, the total annotation number for the whole dataset, and its ground truth generation method. We can see that the VidCrowd dataset has larger resolutions and contains a more average annotation number for each video frame, which is more challenging and suitable for crowd scene analysis and applications.

Some examples of VidCrowd dataset are shown in Fig.~\ref{visual}. Our dataset contains different density levels as shown in the first two rows, which covers a wide range of density varieties. The people number in VidCrowd significantly ranging from 4 to 940. VidCrowd also takes different lighting conditions into considerations with daytime, nightfall, and night video sequences. Besides, our dataset captured in different locations to accommodate different backgrounds, i.e., street, campus, beach, park, squares, etc. Thus VidCrowd is a good candidate for video crowd analysis evaluation.

\begin{table}[!t]
    \centering
    \caption{Ablation study on VidCrowd dataset.}
    \label{tab:ablation}
    \begin{adjustbox}{max width=0.5\textwidth}
			\begin{tabular}{lcc}
				\toprule
				\toprule
				Method                     &MAE    &MSE   \\
				\midrule
				Baseline                   &18.04  &32.86  \\   
				Baseline+non-local         &16.79  &30.93  \\
				\hline
				Motion-guided (ours)       &$\textbf{15.06}$   &$\textbf{29.94}$    \\
				\bottomrule
				\bottomrule
			\end{tabular}
	\end{adjustbox}
\end{table}

\begin{table*}[!t]
    \centering
	\caption{Experiment results of different methods on Mall dataset.}
	\begin{adjustbox}{max width=1\textwidth}
		\label{table:mall}
		\begin{tabular}{lcc}
			\toprule
			\toprule
			Method                     &MAE    &MSE     \\
			\midrule
			Gaussian Process Regression~\cite{chan2008privacy}&3.72   &20.10   \\
			Ridge regression~\cite{chen2012feature}           &3.59   &19.00   \\
			Kernel Ridge Regression~\cite{an2007face}   &3.51   &18.10   \\
			Cumulative Attribute Regression~\cite{chen2013cumulative}&3.43 &17.70 \\
			COUNT forest~\cite{pham2015count}               &2.50   &10.0    \\
			ConvLSTM~\cite{xiong2017spatiotemporal}                   &2.24   &8.50    \\
			Bidirectional ConvLSTM~\cite{xiong2017spatiotemporal}     &2.10   &7.60    \\
			LSTN~\cite{fang2019locality}                       &2.00   &2.50    \\
			E3D~\cite{zou2019enhanced}                        &1.64   &2.13    \\
			\hline
			\name{} (ours)             &$\textbf{1.54}$   &$\textbf{2.02}$ \\
			\bottomrule
			\bottomrule
		\end{tabular}   
		\end{adjustbox}
\end{table*}

\begin{table*}[!t]
    \centering
	\caption{\small Experiment results of different methods on UCSD dataset.}
	\begin{adjustbox}{max width=1\textwidth}
		\label{table:ucsd}
		\begin{tabular}{lcc}
			\toprule
			\toprule
			Method                     &MAE    &MSE   \\
			\midrule
			Ridge Regression~\cite{chen2012feature}           &2.25   &7.82  \\
			Gaussian Process Regression~\cite{chan2008privacy}&2.24   &7.97  \\
			Kernel Ridge Regression~\cite{an2007face}    &2.16   &7.45  \\
			Cumulative Attribute Regression~\cite{chen2013cumulative}&2.07&6.86 \\
			Switch-CNN~\cite{sam2017switching}                 &1.62   &2.10  \\
			Cross-scene~\cite{zhang2015cross}                &1.60   &3.31    \\
			ConvLSTM~\cite{xiong2017spatiotemporal}                   &1.30   &1.79   \\
			\hline
			\name{} (ours)             &\textbf{1.17}   &\textbf{1.45}    \\
			\bottomrule
			\bottomrule
		\end{tabular}  
		\end{adjustbox}
\end{table*}

Our VidCrowd is split into two sets: 6300 frames for training and evaluation, another 2700 frames for testing. The results of our \name{} on VidCrowd compared with other state-of-the-art counting methods are reported in Table~\ref{tab:vidcrowd}. The Mean Absolute Error (MAE) and Mean Squared Error (MSE) are used as evaluation metrics. The results of our \name{} are shown in the last row. And we implement 4 state-of-the-art crowd counting methods: MCNN~\cite{zhang2016single}, CSRNet~\cite{li2018csrnet}, SACANet~\cite{bai2019crowd}, and ConvLSTM~\cite{xiong2017spatiotemporal} as the comparison schemes in our experiment. We observe that our \name{} surpassing all the four methods, which demonstrates the effectiveness of our method.

We conduct an ablation study on VidCrowd to show the importance of \name{} framework. In our \name{}, we use the people flow (motion information) as guidance to promote the density estimation. Besides, \name{} utilize a non-local spatial-temporal network to extract both the local and non-local context information. Without the non-local spatial-temporal module, the network is hard to capture long-range dependencies, thus hinder the counting performance. We compare the results with or without motion guidance and compare the results with or without a non-local spatial-temporal module on VidCrowd, and the results are presented in Table~\ref{tab:ablation}. We can see that the results are further improved with the non-local spatial-temporal module and the motion-guidance, which shows that our \name{} can produce more accurate density maps and promote the counting performance. 
There has been little work on how to leverage the spatial-temporal correlation to improve crowd counting in videos. Monet based on non-local and motion-guided modules to capture both short and long-range contextual information between frames to achieve highly accurate crowd estimation. 
And this newly collected challenging video crowd counting dataset will be released to contribute to the community.

\subsection{Evaluation on Mall and UCSD Benchmarks} \label{mall}

The Mall dataset contains 2000 frames with resolutions $480 \times 640$, which was collected from a public surveillance webcam in a shopping mall. This is a widely used public dataset for video crowd counting evaluations. The region of interest and the perspective map is also provided in this dataset. For a fair comparison, we use the first 800 frames for training and the remaining 1200 frames for testing. We compare our \name{} with other crowd counting algorithms on Mall dataset and some results are shown in Table~\ref{table:mall}. We observe that our approach outperforms all the existing video-based crowd counting algorithms in terms of both MAE and MSE, which demonstrates the effectiveness of our method.

The UCSD dataset consists of 2000 frames of pedestrians on a walkway of the UCSD campus captured by a stationary camera. The video was recorded at 10fps with dimension $238 \times 158$. For a fair comparison, we use the 601-1400 frame for training and the remaining 1200 frames for testing. The experiment results of different methods are shown in Table~\ref{table:ucsd}. We can see that our \name{} is comparable with the state-of-the-art counting algorithms. But this video crowd counting dataset is almost saturated with relatively low-density levels.


\section{Conclusion} \label{sect:conc}

We have proposed \name{}, a novel and highly accurate motion-guided non-local spatial-temporal network for video crowd counting. \name{} not only captures the temporal correlations in video sequences, but also combines both the long-range spatial and temporal contextual information features to boost the counting performance for video data. Besides, we present to the community VidCrowd, a new large-scale challenging video crowd counting dataset offering a diversity of the scene, crowd density, lighting, resolution, etc.
Extensive experiments on VidCrowd, Mall and UCSD datasets show that \name{}  achieves significantly better results as compared with prior arts in terms of MAE and MSE.

\clearpage

{\small
    \bibliography{main}
}

\clearpage
\onecolumn
\appendix

\begin{figure*}[t]
	\centering
	\includegraphics[width=1\textwidth]{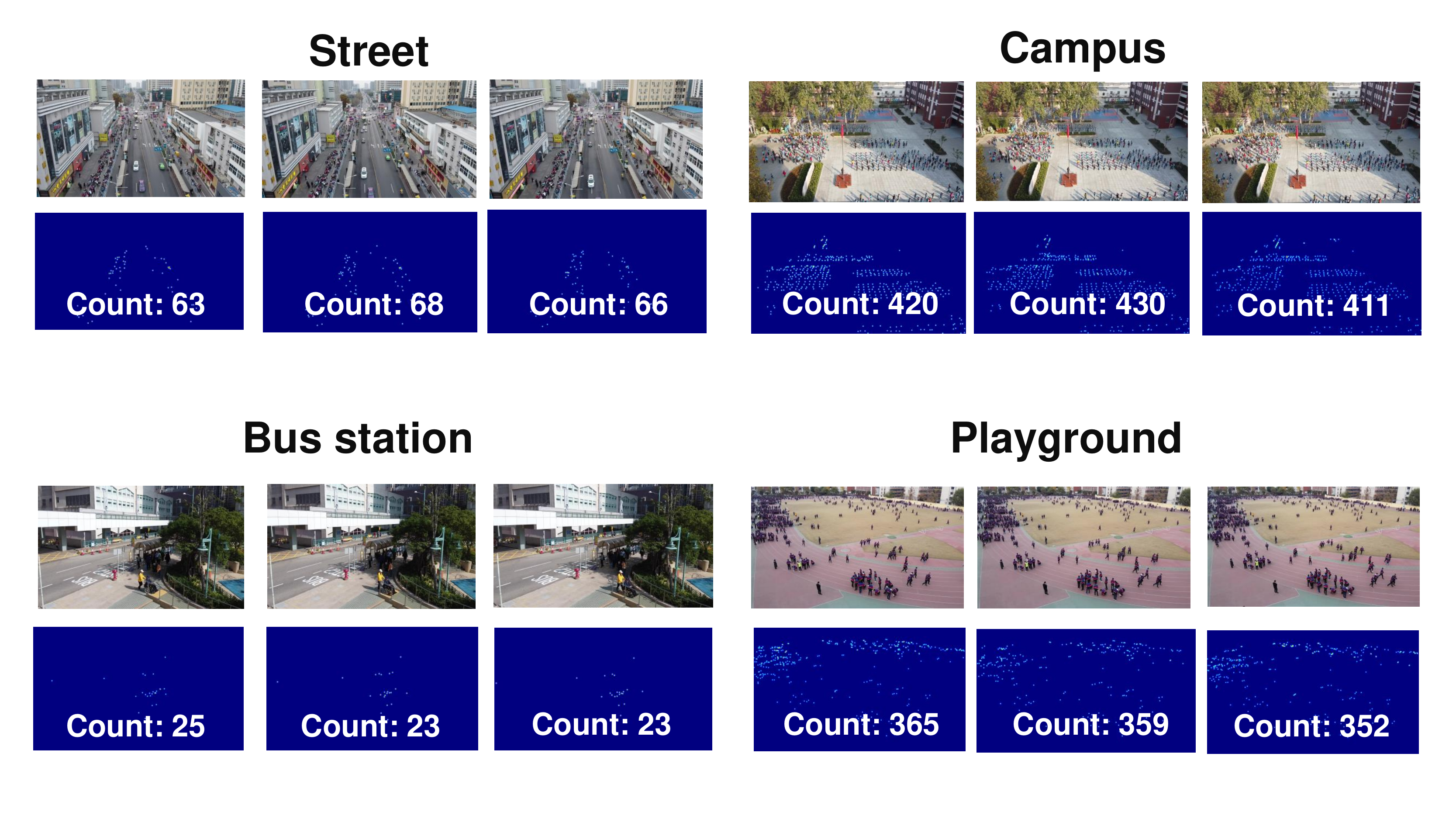}
	\caption{We visualize some examples of the newly collected VidCrowd dataset with its density map, which are captured in different scenes (street, campus, bus station and playground).
	}
\end{figure*}

\begin{figure*}[t]
	\centering
	\includegraphics[width=0.98\textwidth]{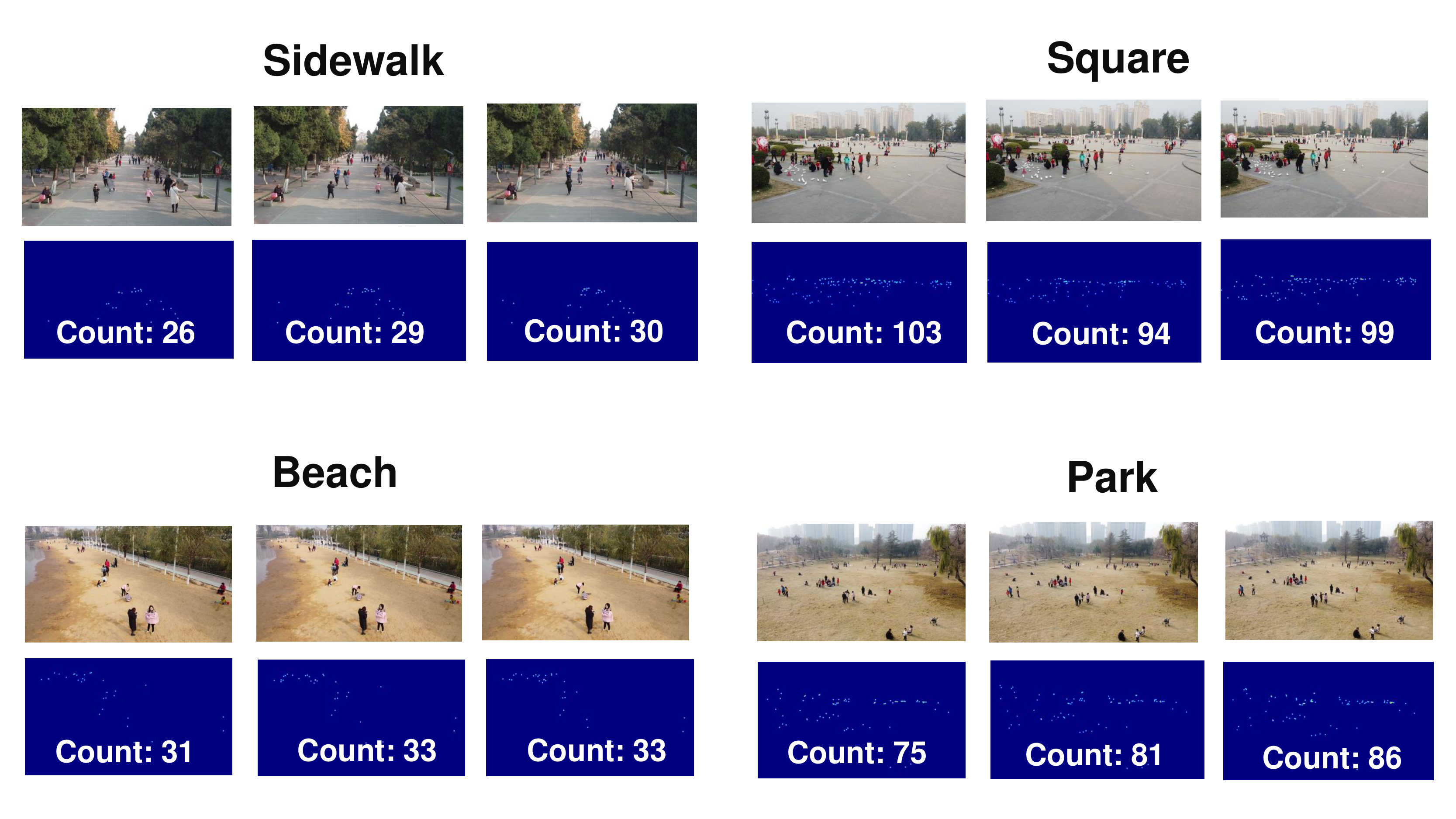}
	\caption{More examples of the VidCrowd dataset with its density map, captured in different scenes (sidewalk, square, beach and park).
	}
\end{figure*}

\end{document}